\documentclass[sn-basic]{sn-jnl}% Basic Springer Nature Reference Style/Chemistry Reference Style

\usepackage{graphicx}%
\usepackage{multirow}%
\usepackage{amsmath,amssymb,amsfonts}%
\usepackage{amsthm}%
\usepackage{mathrsfs}%
\usepackage[title]{appendix}%
\usepackage{xcolor}%
\usepackage{textcomp}%
\usepackage{manyfoot}%
\usepackage{booktabs}%
\usepackage{algorithm}%
\usepackage{algorithmicx}%
\usepackage{algpseudocode}%
\usepackage{listings}%
\usepackage{hyperref}%
\usepackage{caption}%
\usepackage{natbib}
\setcitestyle{numbers,comma,square,compress}

\theoremstyle{thmstyleone}%
\theoremstyle{thmstyletwo}%
\theoremstyle{thmstylethree}%
\begin{document}

\title[Article Title]{An Artificial Neural Network for Image Classification Inspired by Aversive Olfactory Learning Circuits in Caenorhabditis Elegans}
\author[1,2,3]{\fnm{Xuebin} \sur{Wang}}\email{wangxb@mail.bnu.edu.cn}
\author[1,2,3]{\fnm{Chunxiuzi} \sur{Liu}}\email{urnotlcxz@mail.bnu.edu.cn}
\author[4]{\fnm{Meng} \sur{Zhao}}\email{zh\_m@tju.edu.cn}
\author[1,2]{\fnm{Ke} \sur{Zhang}}\email{kezhang@bnu.edu.cn}
\author*[1,2,3]{\fnm{Zengru} \sur{Di}}\email{zdi@bnu.edu.cn}
\author*[1,2]{\fnm{He} \sur{Liu}}\email{heliu@bnu.edu.cn}
\affil[1]{\orgdiv{Department of Systems Science, Faculty of Arts and Sciences}, \orgname{Beijing Normal University},  \orgaddress{\city{Zhuhai}, \postcode{519087}, \country{China}}}
\affil[2]{\orgdiv{International Academic Center of Complex Systems}, \orgname{Beijing Normal University}, \orgaddress{\city{Zhuhai}, \postcode{519087}, \country{China}}}
\affil[3]{\orgdiv{School of Systems Science}, \orgname{Beijing Normal University}, \orgaddress{\city{Beijing}, \postcode{100875}, \country{China}}}
\affil[4]{\orgdiv{School of Computer Science and Engineering}, \orgname{Tianjin University of Technology},  \orgaddress{\city{Tianjin}, \postcode{300384}, \country{China}}}

\abstract{This study introduces an artificial neural network (ANN) for image classification task, inspired by the aversive olfactory learning circuits of the nematode Caenorhabditis elegans (C. elegans). Despite the remarkable performance of ANNs in a variety of tasks, they face challenges such as excessive parameterization, high training costs and limited generalization capabilities. C. elegans, with its simple nervous system comprising only 302 neurons, serves as a paradigm in neurobiological research and is capable of complex behaviors including learning. This research identifies key neural circuits associated with aversive olfactory learning in C. elegans through behavioral experiments and high-throughput gene sequencing, translating them into an image classification ANN architecture. Additionally, two other image classification ANNs with distinct architectures were constructed for comparative performance analysis to highlight the advantages of bio-inspired design. The results indicate that the ANN inspired by the aversive olfactory learning circuits of C. elegans achieves higher accuracy, better consistency and faster convergence rates in image classification task, especially when tackling more complex classification challenges. This study not only showcases the potential of bio-inspired design in enhancing ANN capabilities but also provides a novel perspective and methodology for future ANN design.}

\maketitle
\section{Introduction}\label{sec1}

ANNs are mathematical models inspired by the biological brain, designed to mimic its adaptive learning capabilities. These systems can modify their internal topologies in response to new data, thereby enhancing their performance on a variety of tasks. ANNs have been particularly effective in tackling complex problems such as machine vision \cite{image2016recog,vision2023}, speech recognition \cite{hinton2012speech,speech2023review} and autonomous driving \cite{selfdriving2016}, which were previously challenging for rule-based programming methods. At the same time, machine learning can guide protein engineering efforts \cite{wait2024} and has recently demonstrated effectiveness in designing enzymes \cite{saito2021}, fluorescent proteins\cite{romero2009,wu2019,saito2018}, and optogenetic tools\cite{bedbrook2019,unger2020}.

Despite their remarkable achievements, several challenges remain:
1. High-performing ANNs often contain an excessive number of parameters \cite{para2013}, which can hinder their practical deployment.
2. Training such networks demands substantial computational resources and time, a process that is becoming more costly as the improvements predicted by Moore's Law diminish \cite{res2024limit,moore2017}.
3. The performance of some ANNs is highly scenario-specific, with a significant drop in efficacy when applied to different contexts \cite{sze2017}. 
These issues impact the ANNs' efficiency in terms of size, training speed, and generalization, which are critical considerations for real-world applications.

Biological brains are a rich source of inspiration for the advancement of ANNs \cite{hassabis2017,kriege2015}. Even the most diminutive organisms in the natural world, such as worms, fruit flies and ants, exhibit remarkable capabilities in locomotion and cognitive functions \cite{insects2022,li2020high}. The emulation of cognitive processes and paradigms from biological brains offers a promising avenue to enhance the efficacy of contemporary ANNs \cite{neuroarch2008,biologi2023}. The nematode C. elegans is a paradigmatic model in neurobiological research, notable for its simplicity with a nervous system comprising just 302 neurons \cite{Celegans1986structure}. Despite its simplicity, this primitive brain orchestrates a sophisticated array of behaviors in the worm, including perception, evasion, foraging, learning, and mating. Recent research has shown that an ANN, loosely modeled after the neural structure of C. elegans, can successfully manage the navigation of autonomous vehicles with a significantly reduced network of only 19 neurons \cite{ncp2020,cfctNN2022}. These studies have drawn from the neural information processing mechanisms of C. elegans, but they have not fully replicated the intricate interneuronal connections. Neurons, as the foundational elements of the brain, are organized topologically in a manner that is integral to the functionality and efficiency of the nervous system. Adopting the topological organization observed in the C. elegans nervous system presents a significant opportunity to augment the performance of current ANNs.

The biological neural network of C. elegans, with its 302 neurons and over 7,000 synaptic connections, exhibits a complex topological organization\cite{celeganConn2019}. Through a process of functional clustering, we have effectively distilled this network into a more refined structure comprising 121 functional neurons (as depicted in Fig.\ref{fig1}f). This streamlined model retains the capacity to regulate all vital activities of C. elegans, which are orchestrated by a series of interconnected functional neural circuits. These circuits consist of sensory neurons, interneurons, and motor neurons, with sensory neurons initiating neural information that passes through interneurons before reaching motor neurons to execute actions. Translating the intricate topological structure of C. elegans' neural network into an ANN is challenging due to the unidirectional flow of information in ANNs. However, leveraging a behavioral experiment focused on learning to avoid adverse olfactory stimuli, we have delineated the functional neural circuits responsible for this behavior within the C. elegans neural network. These circuits feature a compact topological structure and a singular direction of neural information transfer, making them an ideal template for the architecture of an ANN. Inspired by the aversive olfactory learning circuits of C. elegans, we have designed an ANN for image classification. This ANN demonstrates superior performance across multiple public datasets when compared to control networks, showcasing the potential of bio-inspired design in enhancing the capabilities of artificial intelligence systems.

This paper is structured as follows: In Sec.\ref{sec2}, we describe a behavioral experiment designed to elicit learning in response to aversive olfactory stimuli. Utilizing high-throughput gene sequencing technology, we identified the functional neural circuits within the complete biological neural network of C. elegans that underpin this learned behavior. Sec.\ref{sec3} details the design of an ANN for image classification, inspired by the topological organization of the functional neural circuits identified in C. elegans. Additionally, we present the design of two other ANNs with distinct architectures for comparative performance analysis. In Sec.\ref{sec4}, we report the results of testing and comparing the performance of these ANNs across various public datasets. The paper concludes with a discussion of our findings in Sec.\ref{sec5}.

\section{Functional neural circuits of aversive olfactory learning in C. elegans}\label{sec2}

\begin{figure}[htbp]
\centering
\includegraphics[width=1.0\textwidth]{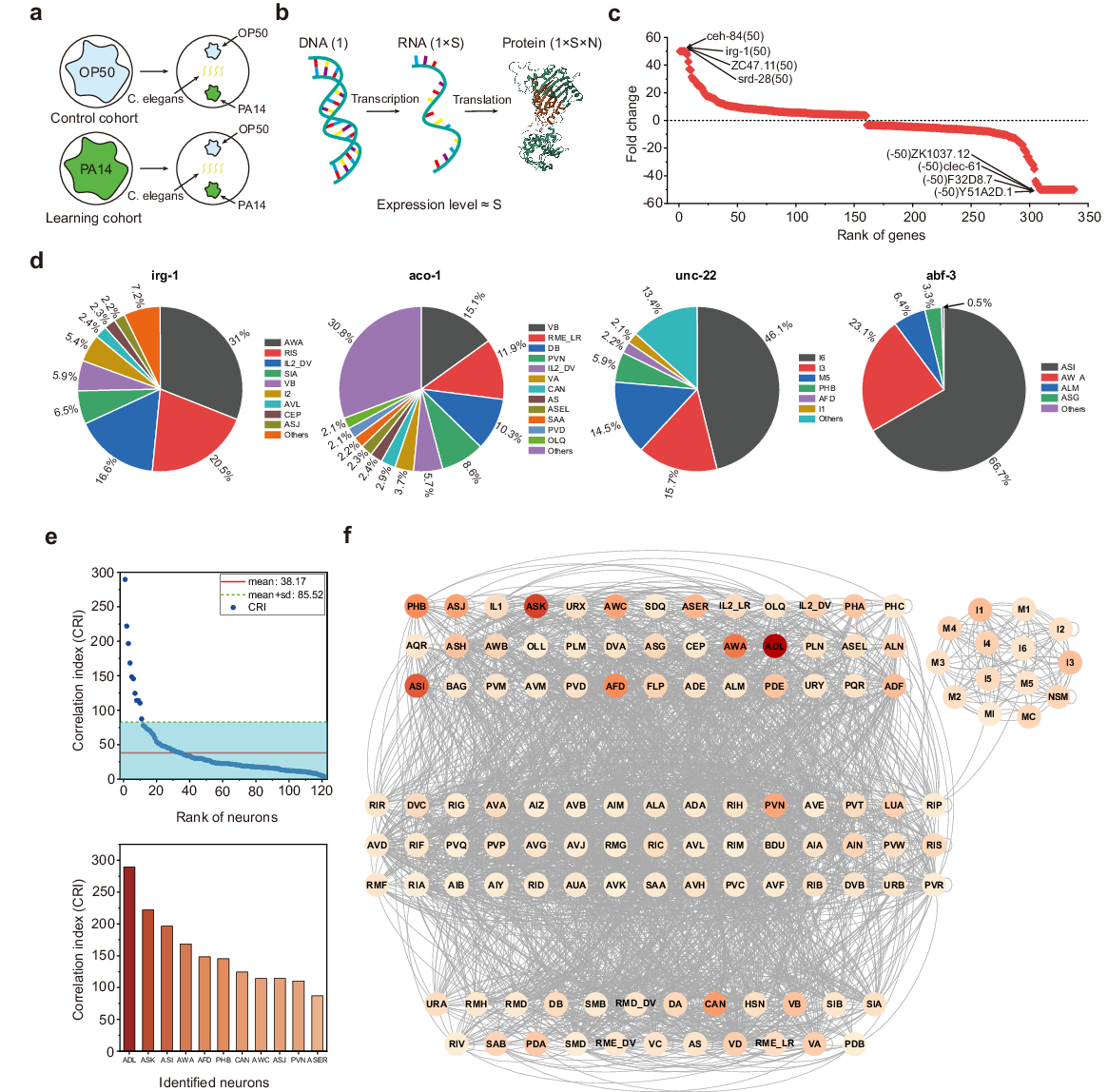}
\caption{
\textbf{a}, Implementation principle of behavioral experiment on learning to avoid aversive olfactory stimuli in C. elegans.
\textbf{b}, Fundamentals of high-throughput gene sequencing technology. 
\textbf{c}, Gene expression fold changes between learning and control cohorts. Positive values denote upregulated and negative values indicate downregulated. To minimize the measurement error, we limit the fold changes to fall within the range of [-50, 50]. 
\textbf{d}, Gene expression percentage in specific functional neurons, excluding functional neurons with expression percentage below 2\%. 
\textbf{e}, Identification of functionally correlated neurons in aversive olfactory learning.
\textbf{f}, The biological neural network of C. elegans. The weight of connection is proportional to the number of synapses $EW_{ij}$ between two functional neurons, while the intensity of color within the functional neurons indicates the magnitude of $CRI_{i}$. 
}
\label{fig1}
\end{figure}

The following procedural steps outline the behavioral experiment designed to assess the learning capabilities of C. elegans in avoiding aversive olfactory stimuli\cite{liu2022forgetting} (as depicted in Fig.\ref{fig1}a). The experiment initiates with two distinct cohorts of C. elegans, both reared under standard laboratory conditions. Each cohort is separately exposed to two types of bacterial strains for a period of 4 hours: the non-pathogenic Escherichia coli OP50 and the pathogenic Pseudomonas aeruginosa PA14 \cite{tPA141999,ha2010,jin2016}. Post-exposure, the worms from each cohort are carefully positioned at the center of a distinct culture plate. These plates are uniformly coated with the OP50 on one half and the PA14 on the other, creating a dichotomy in olfactory stimuli. The critical observation phase involves assessing the distribution of worms as they migrate towards their preferred bacterial strain. Ultimately, the experiment concludes with a quantification of the preference ratios for both OP50 and PA14 among the worms in each cohort. This analysis provides insight into the nematodes' behavioral inclinations. The results of the experiment indicate that C. elegans can develop aversive memories associated with the odor of PA14 after a 4-hour learning period, as evidenced by their preferential avoidance of the pathogenic strain \cite{liu2018cholinergic}.

We aim to elucidate the molecular mechanism of genes expression regulation that
underlies aversive olfactory learning in C. elegans by employing high-throughput gene
sequencing technology (as shown in Fig.\ref{fig1}b). The core concept of this technology is based on the transcription of a single gene in DNA into $S$ RNA molecules, which are then translated into $S \times N$ proteins, with $N$ representing a fixed constant. The level of gene expression is thus represented by $S$. We have quantified the gene expression levels in both the control and learning cohorts as illustrated in Fig.\ref{fig1}a. Comparative analysis of gene expression between the two cohorts has led to the identification of 338 genes exhibiting differential expression, where positive values signify upregulation and negative values denote downregulation, as depicted in Fig.\ref{fig1}c. However, the initial results did not pinpoint the specific neurons accountable for these expression variations. Prior studies have established the expression patterns of these 338 genes across individual neurons in naive C. elegans, with Fig.\ref{fig1}d providing a graphical representation of select examples. A plausible inference is that the 338 genes expression will undergo changes in all neurons expressing it proportionate to its expression in naive C. elegans during aversive olfactory learning. To measure the activity of neurons in relation to aversive olfactory learning, we introduce the correlation index (CRI). The formula for calculating CRI is as follows:
\begin{equation}
\label{eq1}
CRI_{i}={\textstyle\sum_{j=1}^{N}(W_{ji}\times\left|M_{j}\right|)}.
\end{equation}
Here, $CRI_{i}$ represents the correlation index for neuron $i$, $W_{ji}$ is the expression proportion of gene $j$ in neuron $i$ within naive C. elegans, $M_{j}$ is the fold change in gene expression for gene $j$ and $N$ is the total number of genes that displayed differential expression following a 4-hour learning period. The CRIs for 121 functional neurons were computed independently, with the results summarized in Table.\ref{tab1}. Employing statistical analysis on the CRIs of all listed functional neurons from Table.\ref{tab1}, we have identified 11 functional neurons with a robust correlation to aversive olfactory learning in C. elegans, as presented in Fig.\ref{fig1}e. Interestingly, the majority of the 11 identified functional neurons are sensory neurons, with the remainder being one interneuron and one motor neuron.

\begin{table}[htbp]
\caption{Correlation indexes of functional neurons}
\label{tab1}
\begin{tabular}{@{}llllllll@{}}
\toprule
Neuron & CRI & Neuron & CRI & Neuron & CRI & Neuron & CRI\\
\midrule
I1 & 76.20 & SDQ & 15.04 & RIS & 38.09 & RID & 11.17\\
I2 & 24.14 & AQR & 18.17 & ALA & 21.37 & AVB & 11.00\\
I3 & 73.24 & PQR & 21.58 & PVQ & 6.22 & AVA & 33.75\\
I4 & 48.71 & ALM & 16.92 & ADA & 13.50 & PVC & 12.56\\
I5 & 33.67 & AVM & 15.57 & RIF & 18.14 & RIP & 11.22\\
I6 & 19.02 & PVM & 21.70 & BDU & 16.73 & URA & 29.45\\
M1 & 24.11 & PLM & 16.86 & PVR & 10.77 & RME\_LR & 30.14\\
M2 & 27.29 & FLP & 45.12 & AVF & 12.41 & RME\_DV & 17.44\\
M3 & 23.28 & DVA & 22.71 & AVH & 22.14 & RMD\_DV & 17.48\\
M4 & 46.37 & PVD & 23.26 & PVP & 17.54 & RMD & 13.30\\
M5 & 20.83 & ADE & 23.08 & LUA & 47.80 & RIV & 8.79\\
MC & 39.46 & PDE & 66.95 & PVN & 110.22 & RMH & 16.11\\
MI & 11.88 & PHA & 64.58 & AVG & 19.56 & SAB & 42.23\\
NSM & 40.44 & PHB & 145.40 & DVB & 20.38 & SMD & 9.65\\
ASI & 196.67 & PHC & 15.94 & RIB & 26.92 & SMB & 12.98\\
ASJ & 114.12 & IL2\_DV & 51.08 & RIG & 12.16 & SIB & 17.72\\
AWA & 168.32 & IL2\_LR & 35.32 & RMG & 7.73 & SIA & 25.58\\
ASG & 39.42 & CEP & 16.85 & AIB & 5.76 & DA & 37.78\\
AWB & 44.72 & URY & 22.48 & RIC & 31.01 & PDA & 59.73\\
ASEL & 29.98 & OLL & 10.73 & SAA & 15.27 & DB & 34.03\\
ASER & 87.19 & OLQ & 19.05 & AVK & 4.13 & AS & 19.04\\
ADF & 77.80 & IL1 & 32.00 & DVC & 36.16 & PDB & 8.43\\
AFD & 148.39 & AIN & 42.26 & AVJ & 11.69 & VA & 54.20\\
AWC & 114.14 & AIM & 12.75 & PVT & 30.11 & VB & 71.87\\
ASK & 221.88 & RIH & 22.59 & AVD & 15.52 & VD & 52.25\\
ASH & 70.08 & URB & 22.63 & AVL & 12.18 & CAN & 124.36\\
ADL & 289.51 & RIR & 18.92 & PVW & 29.92 & HSN & 27.93\\
BAG & 17.44 & AIY & 10.83 & RIA & 8.25 & VC & 20.47\\
URX & 27.61 & AIA & 22.60 & RIM & 5.32\\
ALN & 47.57 & AUA & 15.91 & AVE & 9.81\\
PLN & 29.92 & AIZ & 9.92 & RMF & 18.69\\
\botrule
\end{tabular}
\end{table}

\begin{figure}[htbp]
\centering
\includegraphics[width=1.0\textwidth]{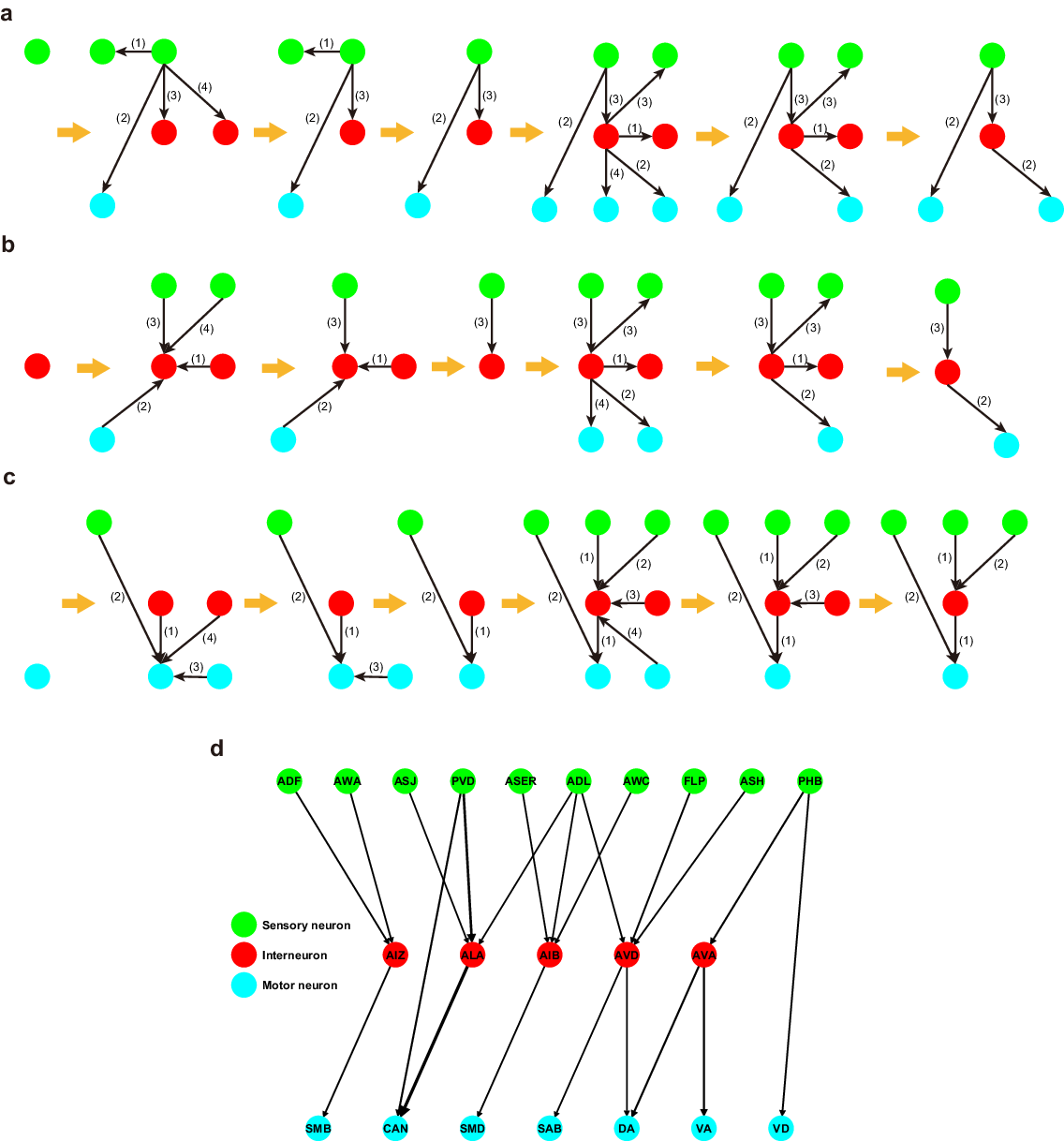}
\caption{
\textbf{a}, The steps of extending the functional neural circuits with sensory neuron as starting neuron. The bracketed value next to arrow line represents the rank determined by the number of synapses $EW_{ij}$ contained in this synaptic connection from large to small. 
\textbf{b}, The steps of extending the functional neural circuits with interneuron as intermediate neuron.
\textbf{c}, The steps of extending the functional neural circuits with motor neuron as terminal neuron.
\textbf{d}, The functional neural circuits associated with aversive olfactory learning of C. elegans,
which consist of 22 functional neurons (10 sensory neurons, 5 interneurons and 7 motor
neurons) and 21 synaptic connections.
}
\label{fig2}
\end{figure}

The neural information flow within the C. elegans brain originates from sensory neurons, subsequently traverses through interneurons and ultimately culminates at motor neurons. Following the sequence of this neural information flow, we have constructed the biological neural network of C. elegans, as depicted in Fig.\ref{fig1}f. In this network representation, sensory neurons are positioned in the upper layer, followed by interneurons in the middle layer and motor neurons in the lower layer. The intensity of the color within each functional neuron correlates with the magnitude of the CRI, providing a visual representation of their respective activities in the context of aversive olfactory learning. The weight of connection $EW_{ij}$ between any two functional neurons is determined by the total number of chemical and electrical synapses present, thus reflecting the strength of communication from neuron $i$ to neuron $j$. This is quantified by the following equation:
\begin{equation}
\label{eq2}
EW_{ij}=C_{ij}+E_{ij},
\end{equation}
In this equation, $EW_{ij}$ represents the weighted connection strength from functional neuron $i$ to $j$. The variables $C_{ij}$ and $E_{ij}$ denote the counts of chemical and electrical synapses, respectively, that originate from neuron $i$ and synapse onto neuron $j$. It is important to note that chemical synapses facilitate unidirectional transmission of neural signals, which is often expressed as $C_{ij}\ne C_{ji}$, while electrical synapses permit bidirectional communication, signified by $E_{ij}=E_{ji}$.

Utilizing the connectivity map of functional neurons as depicted in Fig.\ref{fig1}f, in conjunction with the 11 functional neurons with a robust correlation to aversive olfactory learning filtered by their CRIs as shown in Fig.\ref{fig1}e, we have systematically identified the neural circuits critical for aversive olfactory learning within the biological neural network of C. elegans. The identification process is outlined in the following steps:

\textbf{Scenario One: Sensory Neuron Initiation}

The scenario involves using screened nine sensory neurons as the starting neurons for discovering the functional neural circuits. The three strongest presynaptic connections of each starting neuron are oriented towards second-step neurons. If second-step neurons comprise interneurons or motor neurons, the interneurons or motor neurons and the strong synapses from starting neurons to them will be retained. The strong presynapses of each interneuron in second-step neurons are oriented towards third-step neurons. If third-step neurons comprise motor neurons, the motor neurons and the strong synapses from interneurons in second-step to them will be retained (Fig.\ref{fig2}a).

\textbf{Scenario Two: Interneuron as Intermediary}

The scenario involves using screened one interneuron as the intermediate neuron of the functional neural circuits. The three strongest postsynaptic connections of the interneuron are oriented towards first-step neurons. If first-step neurons comprise sensory neurons, the sensory neurons and the strong synapses from them to the interneuron will be retained. The three strongest presynaptic connections of the interneuron are oriented towards third-step neurons. If third-step neurons comprise motor neurons, the motor neurons and the strong synapses from the interneuron to them will be retained (Fig.\ref{fig2}b).

\textbf{Scenario Three: Motor Neuron as Terminal}

The third scenario involves using screened one motor neuron as the terminal neuron of the functional neural circuits. The three strongest postsynaptic connections of the motor neuron are oriented towards second-step neurons. If second-step neurons comprise sensory neurons or interneurons, the sensory neurons or interneurons and the strong synapses from them to the motor neuron will be retained. The three strongest postsynaptic connections of each interneuron in second-step neurons are oriented towards first-step neurons. If first-step neurons comprise sensory neurons, the sensory neurons and the strong synapses from them to interneurons in second-step neurons will be retained (Fig.\ref{fig2}c). 

Finally, the unretained neurons and synapses are removed to get the functional neural circuits associated with aversive olfactory learning of C. elegans, which consist of 22 functional neurons (10 sensory neurons, 5 interneurons and 7 motor neurons) and 21 synaptic connections (Fig.\ref{fig2}d). Compared to a fully connected neural network with same number of neurons, the sparsity of the functional neural circuits is 96\%.

\section{Artificial neural networks for image classification}\label{sec3}

\begin{figure}[htbp]
\centering
\includegraphics[width=0.83\textwidth]{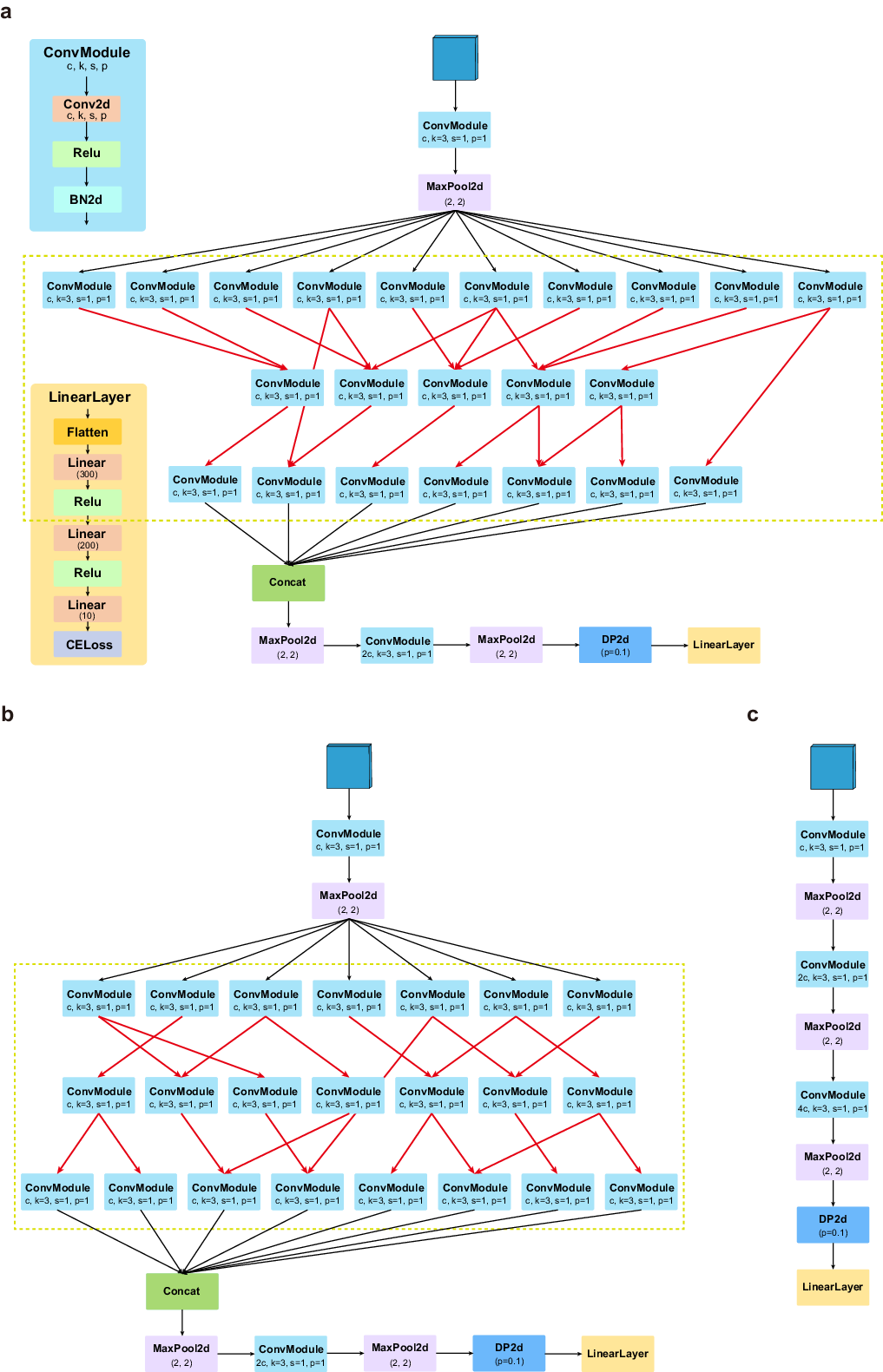}
\caption{
\textbf{a}, The framework map of an artificial neural network for image classification inspired by the neural circuits responsible for aversive olfactory learning in C. elegans. The part framed by green dotted line is the functional neural circuits module of nematode.
\textbf{b}, The framework map of an artificial neural network for image classification inspired by the randomized neural circuits. The part framed by green dotted line is the randomized functional neural circuits module of nematode.
\textbf{c}, The framework map of an artificial neural network for image classification modeled on LeNet \cite{lecunNet1998}.
}
\label{fig3}
\end{figure}

Image classification stands as a central task within the field of computer vision, aiming to differentiate between various categories of images based on their distinctive features. Mathematically framed, it involves the discovery of a function capable of mapping the pixel intensities of an image to a specific class label. Humans, endowed with extensive prior knowledge, can effortlessly categorize images subconsciously. In contrast, for computers, discerning image categories from mere pixel values is a non-trivial challenge. Prior to the deep learning revolution, image classification often relied on machine learning models that utilized manually crafted features, a process heavily dependent on expertise and iterative experimentation. The emergence of deep learning has significantly alleviated the need for manual feature engineering. Artificial neural networks, particularly those with a vast number of parameters, are now capable of autonomously extracting features and classifying images. Convolutional neural networks (CNNs)\cite{fukushima1980}, which form the backbone of most contemporary image classification models, draw inspiration from the mechanisms of human visual processing. Unlike traditional fully connected neural networks, CNNs employ convolution operations to capture local features within the input data. Through training, these networks are able to automatically optimize the parameters of the convolutions applied. By stacking multiple CNN layers and refining their interplay, the precision of image classification can be significantly enhanced. Sequential combinatorial architectures, such as the foundational LeNet, are among the most widely adopted CNN frameworks for image classification.

Identifying the optimal assembly of modules to construct an ANN for image classification is a challenging task. The neural circuits of the C. elegans have undergone evolutionary optimization, exhibiting a topological organization that enhances the processing of neural information. Drawing inspiration from the functional neural circuit's topological organization illustrated in Fig.\ref{fig2}d, we have designed an ANN tailored for image classification, as presented in Fig.\ref{fig3}a. To highlight the benefits of the topological design depicted in Fig.\ref{fig3}a, we constructed a comparative network Fig.\ref{fig3}b. This comparison ANN features a randomized functional neural circuits module. Notably, this randomization is performed while maintaining an equal number of functional neurons and synaptic connections to those found in the original network. Furthermore, as a second point of comparison, we developed a sequential combinatorial ANN for image classification Fig.\ref{fig3}c. This network was designed by referencing the foundational architecture of LeNet, a well-established image classification CNN model.

The topological organization inherent in the functional neural circuits module is intricate and substantially divergent from the linear, sequential combinations commonly found in traditional neural network architectures. This complexity, as seen in the networks presented in Fig.\ref{fig3}a and Fig.\ref{fig3}b, is postulated to enhance the potential for improved accuracy in image classification tasks. The intricate interplay of connections within the network allows for a more nuanced and potentially powerful representation of the data, which can be crucial for the classification of images. Moreover, the three varieties of artificial neural networks for image classification showcased in Fig.\ref{fig3} offer the flexibility to produce models with varying numbers of parameters. This flexibility is achieved by modulating the count of convolutional kernels denoted by $c$. It is well-established that models with different parameter counts, which directly influence the capacity of the network, can exhibit divergent levels of accuracy when classifying images. The balance between the number of parameters and the complexity of the model is a critical factor in achieving optimal performance, as it affects the network's ability to generalize from training data to unseen images without overfitting.

\section{Results}\label{sec4}

\begin{table}[htbp]
\caption{Public image datasets for classification}
\label{tab2}
\begin{tabular}{@{}llllll@{}}
\toprule
Dataset & Color channel & Size & Category & Training set & Test set\\
\midrule
MNIST\cite{lecunNet1998} & 1 & 28x28 & 10 & 60000 & 10000\\
FashionMNIST\cite{fashionMnist2017} & 1 & 28x28 & 10 & 60000 & 10000\\
CIFAR10\cite{cifar102009} & 3 & 32x32 & 10 & 50000 & 10000\\
CIFAR100\cite{cifar102009} & 3 & 32x32 & 100 & 50000 & 10000\\
\botrule
\end{tabular}
\end{table}

\begin{figure}[htbp]
\centering
\includegraphics[width=1.0\textwidth]{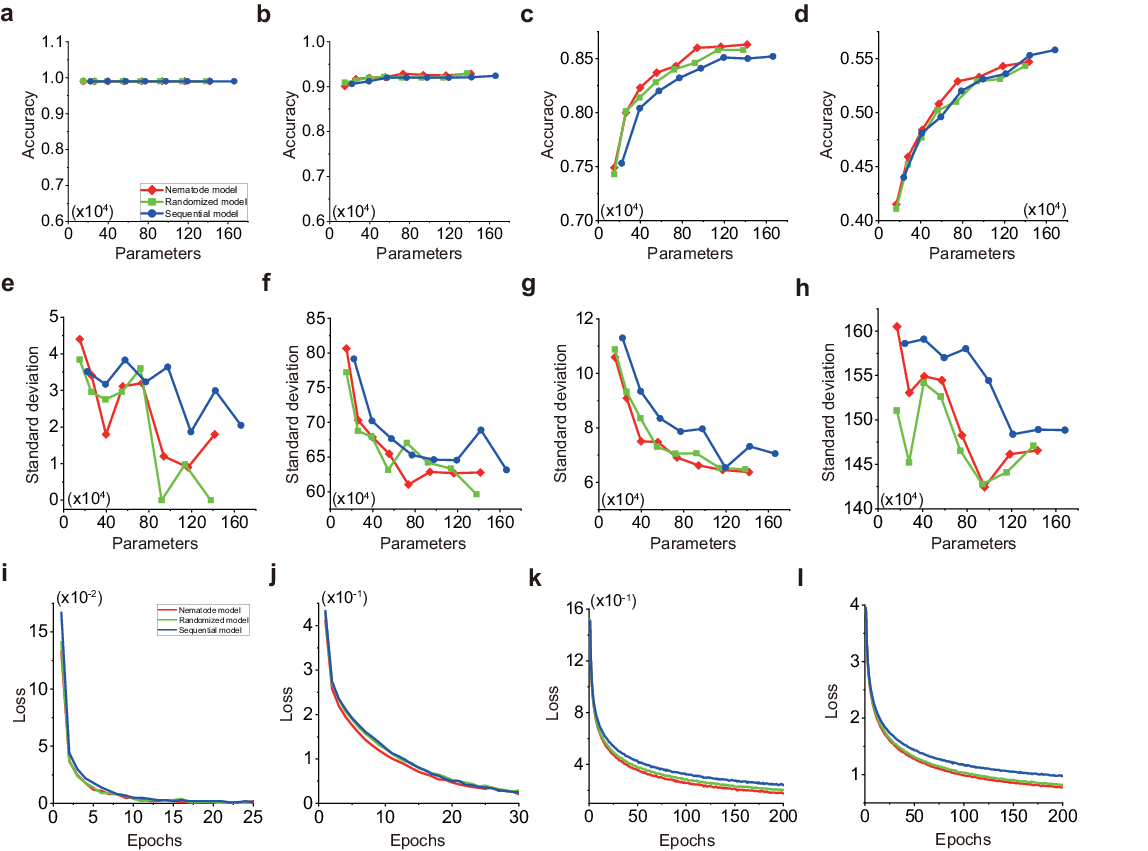}
\caption{
The accuracy of classification across all categories on MNIST (\textbf{a}), FashionMNIST (\textbf{b}), CIFAR10 (\textbf{c}), CIFAR100 (\textbf{d}). The consistency of classification accuracy among different categories on MNIST (\textbf{e}), FashionMNIST (\textbf{f}), CIFAR10 (\textbf{g}), CIFAR100 (\textbf{h}). The rate of convergence of classification loss across all categories on MNIST (\textbf{i}), FashionMNIST (\textbf{j}), CIFAR10 (\textbf{k}), CIFAR100 (\textbf{l}). 
}
\label{fig4}
\end{figure}

Our models underwent rigorous validation using a multitude of public datasets, as enumerated in Table.\ref{tab2}. This validation process was designed to assess three critical aspects of performance:
\begin{enumerate}
\item The overall accuracy of classification across all represented categories.
\item The uniformity or consistency in classification accuracy among the various categories.
\item The rate at which the classification loss converges across all categories, indicative of the model's learning efficiency.
\end{enumerate}

Fig.\ref{fig4}(a-d) present the classification accuracy of three distinct ANN models for image classification across various datasets. On simpler datasets such as MNIST and FashionMNIST, the image classification ANN models, regardless of their architectures, demonstrate comparable accuracies. This pattern shifts with more complex datasets like CIFAR-10 and CIFAR-100, where image classification ANN models drawing inspiration from the neural circuits of aversive olfactory learning in C. elegans excel. These are closely followed by image classification models based on randomized neural circuits, both surpassing the performance of sequential image classification ANN models akin to LeNet in terms of accuracy. Fig.\ref{fig4}(e-h) examines the classification consistency across different categories for these models. Here, the sequential image classification models exhibit the least consistency, whereas the image classification models inspired by nematode neural circuits and those inspired randomized neural circuits show comparable levels of consistency. Additionally, the convergence speed, as shown in Fig.\ref{fig4}(i-l), is consistent with the pattern of classification accuracy.

The outcomes, as depicted in Fig.\ref{fig4}, reveal that the ANN for image classification, which draws inspiration from the aversive olfactory learning circuits of C. elegans, exhibits a notable advantage. This advantage becomes particularly pronounced when the models are tasked with more complex image classification challenges.

\section{Conclusion and discussion}\label{sec5}

The research presented in this paper explores the potential of bio-inspired design in the development of ANNs for image classification. The study draws inspiration from the C. elegans, a biological model with a simple yet effective nervous system. Despite the simplicity of C. elegans, with only 302 neurons, it exhibits complex behaviors that are orchestrated by a sophisticated array of neural circuits. This research aimed to replicate the topological organization of these circuits in an ANN to address the challenges faced by conventional ANNs, such as excessive parameterization and high computational training costs.

Through a series of behavioral experiments and high-throughput gene sequencing, the study identified the functional neural circuits within C. elegans that are critical for aversive olfactory learning. These circuits were then translated into an image classification ANN architecture, which was tested and compared with two other image classification ANNs of different architectures across various public image datasets.The results of the study are promising, demonstrating that the ANN inspired by C. elegans' neural circuits outperforms the control networks in terms of classification accuracy, classification onsistency and learning efficiency. The bio-inspired ANN showed a significant advantage, particularly when dealing with more complex image classification tasks. This suggests that the topological organization of the C. elegans neural circuits enhances the network's ability to generalize and learn from data, leading to improved performance.

This research provides evidence that bio-inspired design can significantly enhance the capabilities of ANNs for image classification. The findings suggest that the topological organization observed in the C. elegans nervous system can be effectively translated into ANN architecture, leading to improved performance in complex tasks. This study opens up new avenues for ANN design, encouraging further exploration of biological systems for inspiration in developing more efficient and effective artificial intelligence systems.

\section{Acknowledgements}\label{sec6}

This work is supported by the National Natural Science Foundation of China (NSFC) under grant numbers 32371079 and 32000720.

\bibliography{sn-bibliography}
\end{document}